\pdfoutput=1

\documentclass[11pt]{article}

\usepackage{acl}

\usepackage{times}
\usepackage{latexsym}

\usepackage{color}
\usepackage{amsmath}
\usepackage{bm}
\usepackage{booktabs}
\usepackage{multirow}
\usepackage{array}
\usepackage{paralist}

\usepackage{graphicx}
\usepackage{mathtools, cuted}
\usepackage{dsfont}
\usepackage{hyperref}
\usepackage{amssymb}
\usepackage{amsfonts}
\usepackage{subcaption}
\usepackage{diagbox}
\usepackage[ruled,vlined,noresetcount]{algorithm2e}

\usepackage{xcolor}

\SetCommentSty{mycommfont}
\usepackage[T1]{fontenc}

\usepackage[utf8]{inputenc}

\usepackage{microtype}
\definecolor{timid-blue}{RGB}{221, 235, 247}

\DeclareMathOperator*{\argmax}{arg\,max}
\DeclareMathOperator*{\E}{\mathbb{E}}
\DeclareMathOperator*{\R}{\mathbb{R}}
\DeclareMathOperator*{\Prob}{\mathbb{P}}
\newenvironment{myfont}{\fontfamily{pcr}\selectfont}{\par}
\DeclareTextFontCommand{\textmyfont}{\myfont}
\usepackage{pifont}
\newcommand{\cmark}{\ding{51}}
\newcommand{\xmark}{\ding{55}}
\newcommand\blfootnote[1]{%
  \begingroup
  \renewcommand\thefootnote{}\footnote{#1}%
  \addtocounter{footnote}{-1}%
  \endgroup
}
\title{CRUSH: \underline{C}ontextually \underline{R}egularized and \underline{U}ser anchored \underline{S}elf-supervised \underline{H}ate speech Detection}

\author{Souvic Chakraborty$^{*\dagger}$, Parag Dutta$^{1*\#}$, Sumegh Roychowdhury$^{*\dagger}$, Animesh Mukherjee$^{\dagger}$ \\
    $^\#$Indian Institute of Science (IISc), Bangalore, KA, IN - 560012 \\
    \texttt{paragdutta@iisc.ac.in} \\
    $^{\dagger}$Indian Institute of Technology (IIT), Kharagpur, WB, IN - 721302 \\
    \texttt{\{chakra.souvic,sumeghtech,animeshm\}@gmail.com} \\
  }

\begin{document}

\maketitle

\setlength{\abovedisplayskip}{4pt}
\setlength{\belowdisplayskip}{4pt}
\setlength{\belowcaptionskip}{-15pt}

\begin{abstract}

    The last decade has witnessed a surge in the interaction of people through social networking platforms. While there are several positive aspects of these social platforms, their proliferation has led them to become the breeding ground for cyber-bullying and hate speech. Recent advances in NLP have often been used to mitigate the spread of such hateful content. 
    Since the task of hate speech detection is usually applicable in the context of social networks, we introduce \textit{CRUSH}, a framework for hate speech detection using User Anchored self-supervision and contextual regularization.
    Our proposed approach secures $\approx1$-$12\%$ improvement in test set metrics over best performing previous approaches on two types of tasks and multiple popular English language social networking datasets.
    \blfootnote{\hspace{-0.12cm}$^*$Equal Contribution.}\footnotetext[1]{Work done while at IIT Kharagpur.}
    \blfootnote{Accepted in Findings of NAACL-HLT 2022.}
    \\
    \textit{\textbf{Note}: This paper contains materials that may be offensive or upsetting to some people.}

\end{abstract}


\section{Introduction}
Today, the world is more connected than ever in the history of mankind. This can primarily be attributed to: (i) the technological advancements that have made affordable internet connections and devices available to people, and (ii) the social networking platforms that have hosted and connected these people. As a result, even people divided by geography can seamlessly interact in real-time without stepping outside their homes. In fact, social networks are an integral part of today's society.

We, however, are more concerned about the pitfalls of this global widespread use of social networks. The unprecedented and rapid explosion in social networking and social media use has left many people –- particularly the youth, women, and those from ethnic and religious minority groups –- vulnerable to the negative aspects of the online world like sexual harassment, fake news and hate speech \cite{cyberbullying}.
The number of toxic users dumping their radically biased views and polarising content onto these networks have burgeoned to such a level that they are causing political discord and communal disharmony. Therefore such posts must be intervened upon and filtered before they have the intended effect on the mass. With a huge number of posts on many popular social networking websites every second, manual filtering for hate speech does not scale. Hence, automating hate speech detection has been the primary focus of many researchers in recent years, both in academia and industry alike.

\begin{figure}[t!]
    \centering
    \includegraphics[width=0.8\linewidth]{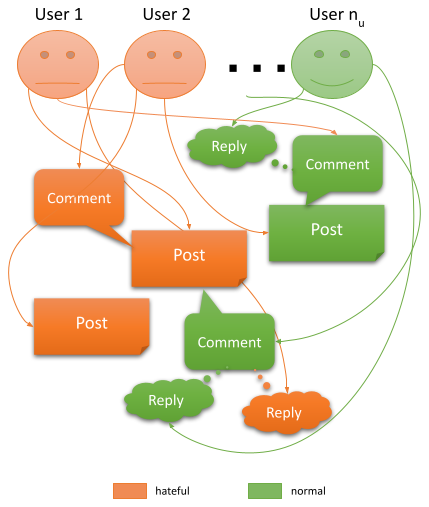}
    \caption{
        [Best viewed in color] {Hateful content tend to cluster together in common threads and usually come from few hateful users in social media. We stress on this informed assumption to learn better representations using self supervision and contextual regularization. In the sub-graph shown in the pic, the textual content is in the form of posts, comments on the posts (optional), and replies to the comments (optional). $n_u$ is the total number of users in the network (dynamic), and each of the text sequences can be attributed to one of the users.}}
    \label{img:SocialGraph}
\end{figure}

Figure \ref{img:SocialGraph} depicts and describes a sub-graph of a typical social network. It is essential to leverage this structure within social networks for infusing network context while identifying hate speech. We investigate and notice that the social graph context can be disentangled into two components: (i) \textit{Post context}: the context in the neighborhood around the text sequence, i.e., the sub-graph consisting of posts, comments, and replies (see Figure \ref{img:SocialGraph}) and (ii) \textit{User context}: the context from all the existing text sequences in the network that originated from the same user (see, for instance, the connections emanating from users 1, 2, $n_u$ etc. in the Figure \ref{img:SocialGraph}). Relying on the echo-chamber effect, we accordingly propose a framework that uses self supervision from unlabelled data harnessed from the social networks (Gab \& Reddit in our case), so that we can use contextual information (user \& post context) to generate better representations from input sentences for hate speech related tasks. The main contributions of this paper are:
\begin{compactitem}
\item [(i)] First, we propose \textbf{UA (User Anchored self-supervision)}, a self-supervised contrastive pre-training objective. Essentially we try to incorporate the mapping from text sequences to users into the language model. In addition, we provide a Robust UA strategy that incorporates the hate speech downstream task information into our proposed UA pre-training approach.
\item [(ii)] Next, we propose \textbf{CR (Contextual Regularization)}, a regularization strategy based on the findings of \citet{10.1145/3415163}. 
Here we introduce a loss based on the informed assumption that the neighboring comments and replies (in the social graph) of a hateful comment is more likely to be hateful than the comments/replies in the vicinity of a non-hateful comment. It helps us to regularize the supervised learning paradigm by learning better representations of text sequences in social network context.
\item[(iii)] We experiment with two types of hate speech tasks -- classification and scoring -- across three datasets. We show that our approach secures $\approx1$-$4\%$ improvement in F1-scores (for classification tasks) and $\approx12\%$ improvement in mean absolute error (for scoring task) when compared to best competing baselines.
\end{compactitem}
To the best of our knowledge, we are the first to use text sequence based self-supervision in hate speech using user characteristics. {One of the key technical contribution that this paper makes is to show that it is more advantageous to use context to regularize a classification model than directly infusing the context into the model.} Also, none of our proposed approaches require any additional annotation effort or introduce any extra parameter into the training pipeline, and are therefore scalable.



\section{Related work}
\label{Section:Related work}

Hate speech is heavily reliant on linguistic complexity. \citet{waseem-2016-racist} showed that classification consensus is rare for certain text sequences even among human annotators. Automatic detection of hate speech is further strongly tied to the developments in machine learning based methods.

Until recently, feature engineering was one of the popularly used techniques. \citet{Gitari2015ALA} designed several sentiment features and \citet{delvigna2017} used the sentimental value of words as the main feature to measure the hate constituted within a text sequence.
Empirical evidence was provided by \citet{malmasi-zampieri-2017-detecting} indicating that n-gram features and sentiment features can be successfully applied to detect hate speech. \citet{rodriguez-2019} constructed a dataset of hate speech from Facebook, and consequently proposed a rich set of sentiment features, including negative sentiment words and negative sentiment symbols, for detecting hateful text sequences.
As witnessed by the above works, it was widely believed that sentiment played an important role in detecting underlying hate. 

More recently, deep learning based methods \cite{10.1145/3041021.3054223} have garnered considerable success in detecting hate speech since they can extract the latent semantic features of text and provide the most direct cues for detecting hate speech. \citet{zhang-2018} developed a CNN+GRU based model to learn higher-level features.
{\cite{kshirsagar-2018} passed pre-trained word embeddings through a fully connected layer}, which achieved better performance than contemporary approaches despite its simplicity. 
\citet{tekiroglu-etal-2020-generating} constructed a large-scale dataset for hate speech and its responses and used the pre-trained language models (LM) for detection. These methods demonstrated the considerable advantages of interpreting words in the context of given sentences over static sentiment based methods.

Self-supervision and auxiliary supervision have also been explored for hate speech detection. HateBERT \cite{caselli-etal-2021-hatebert} used the Masked Language Modelling (MLM) objective to learn contextual hate semantics within text sequences from individual Reddit posts. HateXplain \cite{Mathew_Saha_Yimam_Biemann_Goyal_Mukherjee_2021} used human annotators to obtain rationale about the text containing hate speech and then applied the same for improving their language model. {Researchers have previously tried context infusion in the inputs \cite{menini2021abuse, vidgen-etal-2021-introducing, pavlopoulos2020toxicity}. \citet{deltredici} showed that user context helps in downstream tasks. On the other hand, \citet{menini2021abuse} and \citet{vidgen-etal-2021-introducing} show that contextual classification is harder given contextual annotations. A similar theme recurs when \citet{pavlopoulos2020toxicity} shows that context infusion does not easily increase the performance of classifiers in context of toxicity detection in text. Alternatively instead of using context directly as input we use it as a regularizer to improve the classifier by learning better contextual representations while training. During inference, we use only the post text, thus adding no computation overhead.
}


\begin{figure*}[ht]
    \centering
    \includegraphics[width=0.9\textwidth]{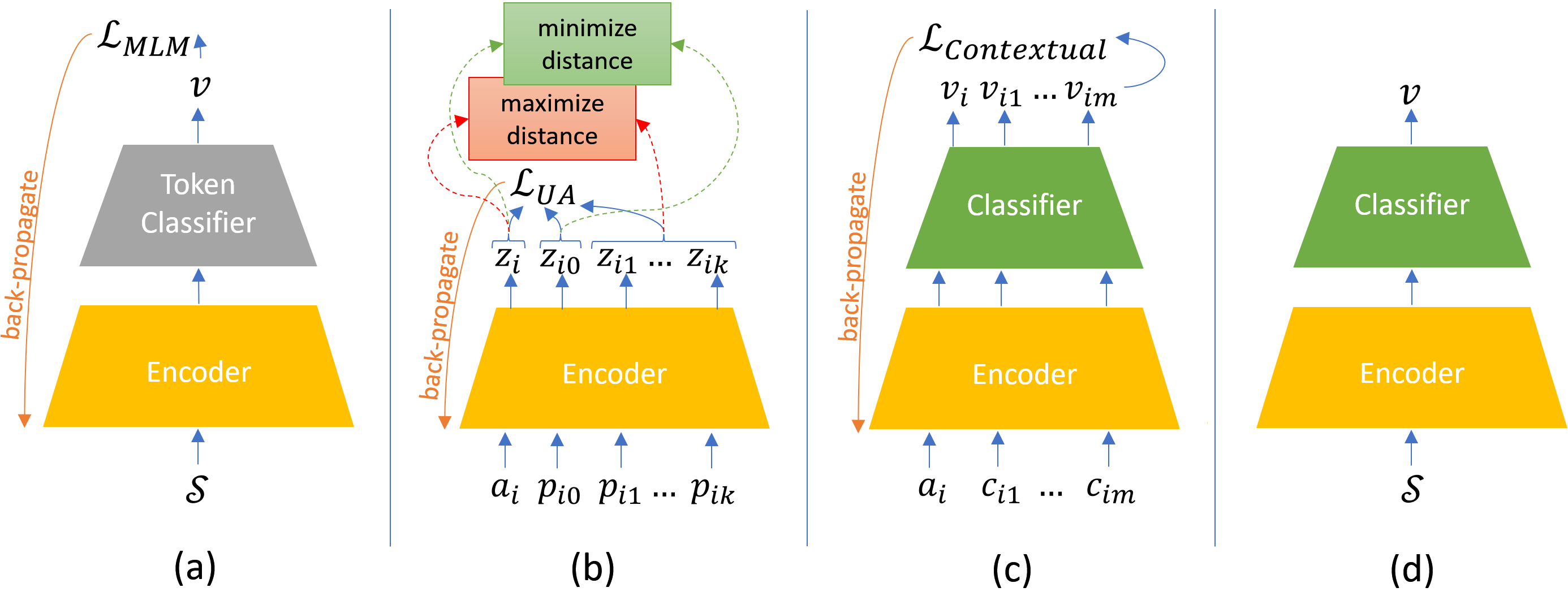}
    \caption{
    [Best viewed in color] An illustration of the various phases of training and inference for the classification task. The regression task will have a similar structure except for a regressor head instead of a classifier head in \textit{(c)} and \textit{(d)}. The blue arrows indicate the forward pass, and the orange arrows indicate the backward pass. \textit{(a)} corresponds to Phase I, i.e. continual pre-training using self-supervised MLM objective on hateful sequences to incorporate contextual hate understanding. \textit{(b)} corresponds to Phase II, i.e. the User Anchored self-supervised learning objective, it depicts the scalable contrastive objective model that does not add any additional parameters. \textit{(c)} corresponds to Phase III, i.e. our contextual regularization procedure where we add post and user context. And finally \textit{(d)} shows the inference phase which does not require any additional context.}
    \label{img:ModelDiagrams}
\end{figure*}

We find that using self-supervision for hate speech detection systems leveraging the associated context within a social network is heavily under-explored. s demonstrated by \citet{10.1145/3415163}, learning in a broader socio-personal context of the users and contemporary social situations is also very important. The authors showed the clustering tendency of hateful text sequences on specific hateful user timelines and after specific events (temporal clustering). However, to the best of our knowledge, no prior work in hate speech detection has explored self-supervision in these directions using the context information.



\section{Proposed approaches}
\label{Section:Proposed methods}
Assumptions about invariance(s) are required in all self-supervised learning objectives. We make the following two assumptions articulated below.
\begin{compactitem}
\item [(i)] In the masked language modeling (MLM) objective, the invariant is that the conditional probability distribution of the vocabulary tokens (or words) for each masked token can be reasonably estimated given the context (in the form of tokens) around the masked token.

\item [(ii)] In the User Anchored self-supervision (UA) objective, we assume that the users' writing style and bias (specifically cultural and in-group bias) are invariant \cite{Hughes7682}. Hence, the inverse mapping of a post to the corresponding user should be estimable subject to the language understanding capability.
\end{compactitem}
We denote the model being trained for the downstream tasks as $\mathcal{M}$. $\mathcal{M}$ has two modules: \textbf{(a)} an encoder for encoding the input sentences, and \textbf{(b)} a classifier or regressor for mapping the hidden representations generated by the encoder into one of $\mathcal{K}$ classes and a single value respectively.
For instance, the encoder in $\mathcal{M}$ can be modeled by a transformer \cite{NIPS2017_3f5ee243}. See Figure \ref{img:ModelDiagrams} for block diagram of our proposed approaches.


\subsection{\textit{Phase I}: Incorporating hateful content into the language model through {continual} pre-training (CP)}
\label{Subsection:Phase1}

We start with pre-trained language models and continue pre-training them by minimizing the standard MLM objective using text sequences (posts) accumulated from a variety of social network datasets. The procedure for domain adaptation in LMs can be found in \citet{gururangan-etal-2020-dont} and a procedure tailored for hateful words can be found in HateBERT. We use the text sequences available on Pushshift\footnote[2]{\hyperlink{https://files.pushshift.io/gab}{https://files.pushshift.io/gab}} in addition to RAL-E from \hyperlink{https://osf.io/tbd58/?view_onlycb79b3228d4248ddb875eb1803525ad8}{HateBERT} for pre-training our Hate-infused Language Model (HateLM). 


\subsection{\textit{Phase II}: User Anchored self-supervision (UA)}
\label{Subsection:Phase2}

Some users are more biased than others and hence are more prone to post toxic content \cite{10.1145/3415163}. We employ a User Anchored self-supervision (UA) objective in the next pre-training phase to compare the users' writing style. Hence, given an LM is capable of language understanding, it should also be able to distinguish between users' writing styles from their posts with high probability, when the pool of posts from various users is not very large.

Here we use a self-supervised method based on \textit{contrastive pre-training} with negative samples to efficiently incorporate UA into an LM. This negative sampling makes the approach highly scalable\footnote[3]{Owing to the extremely large number of users in a platform and the dynamic nature of the graph, it is infeasible to simply add a user classification network after the encoder}.

For the text sequence $\mathcal{S}$ corresponding to the $i^{\text{th}}$ sampled post and user $u_0$, we try to decrease some distance metric between representation of $\mathcal{S}$ and another sentence by $u_0$ among the $n_u$ available users while increasing the distance metric with sentences authored by other users in a contrastive fashion. Next, we sample $k << n_u$ number of users uniformly at random, without repetition, from among the list of all users in the social network. Adding $u_0$ along with the $k$ sampled users in a set, we get:
\begin{align}
    \label{eq:SampledUserSet}
    U_i = \{ u_{i0}, u_{i1}, ... , u_{ik} | \forall_{j_1,j_2 \in \{0,..,k\}}u_{ij_1} \neq u_{ij_2} \}
\end{align}
The set of posts $P_i$ made by users in $U_i$.
\begin{align}
    \label{eq:SampledPostSet}
    P_i = \{ p_{i0}, p_{i1}, ... , p_{ik} | p_{i0} \neq \mathcal{S} \}
\end{align}
$p_{i0}$ cannot be same as $\mathcal{S}$ because our positive sample needs to be different from the original post. The remaining $\forall_{j \in [k]}\{p_{ij}\}$ become negative samples. 




For Phase II pre-training and updating the parameters $\theta_\mathcal{E}$ of the encoder in $\mathcal{M}$, we start by sampling an anchor sequence $a_i \equiv \mathcal{S}$ from our dataset. We pass $a_i$ through encoder in $\mathcal{M}$ to obtain $z_i$.
\begin{align}
    \label{eq:AnchorEmbedding}
    z_i = \mathcal{F}_{\theta_\mathcal{E}}(a_i)
\end{align}
Next, we generate $P_i$ by sampling as described above. We pass all of the $p_{ij}\text{'s} \in P$ through our encoder as well, to get embeddings:
\begin{align}
    \label{eq:SampledSetEmbeddings}
    Z_i = \{z_{i0}, z_{i1}, ..., z_{ik}  \}
\end{align}
We use a self-supervised contrastive loss inspired from \citet{NEURIPS2020_d89a66c7} and modify it for our purposes:
\begin{align}
    \label{eq:UserAttributionContrastiveLoss}
    \mathcal{L}_{\text{UA}} = \E_{i \sim \mathcal{U}([N])} \left[ - log \left( \frac{\textmyfont{exp}(z_i \cdot z_{i0})}{\sum_{j \in [k]} \textmyfont{exp}(z_i \cdot z_{ij})} \right) \right]
\end{align}
Here, `$\cdot$' represents the inner product and $\mathcal{U}(.)$ represents discrete uniform distribution. Since there are no additional models, only $\theta_\mathcal{E}$ parameters need to be updated during backpropagation.

\vspace{-0.2cm}
\paragraph{\textit{Robust UA}:}
During training UA objective exhibits the problem of over-fitting. After a certain number of epochs, the information in the language model accumulated during the CP phase gets overshadowed by information obtained from the UA phase. In order to handle the problem, the model needs to be fine-tuned after every epoch. Since this is not feasible, we propose a method for a more robust training method aligned to our downstream task of hate speech classification. Since we already have the annotations in the target dataset, we add an auxiliary task to align the model parameter updates towards the downstream task rather than diverge as a consequence of over-fitting.

For an anchor sequence $a_i \equiv \mathcal{S}$ sampled from the training set, we get the original class label of the example and get a positive example $p$ by sampling a new post from the training set belonging to the same class. Similarly, we sample two posts from each of the remaining classes in order to get a set of $l = 2(n_c-1)$ negative examples, where $n_c$ is the number of classes in the downstream classification task. Hence similar to $P_i$ in Equation \ref{eq:SampledPostSet}, our auxiliary post set corresponding to the $i^\text{th}$ sampled sequence is as follows:
\begin{align}
    \label{eq:SampledAuxPostSet}
    \overline{P}_i = \{ \overline{p}_{i0}, \overline{p}_{i1}, ... , \overline{p}_{il} | \overline{p}_{i0} \neq \mathcal{S} \}
\end{align}
Subsequently, we pass them through the encoder in $\mathcal{M}$ and generate auxiliary embedding set:
\begin{align}
    \label{eq:SampledAuxSetEmbeddings}
    \overline{Z}_i = \{\overline{z}_{i0}, \overline{z}_{i1}, ..., \overline{z}_{il} \}
\end{align}
Similar to the supervised contrastive loss described in Equation \ref{eq:UserAttributionContrastiveLoss}, we add the following auxiliary loss:
\begin{align}
    \label{eq:AuxContrastiveLoss}
    \mathcal{L}_{\text{aux}} = \E_{i \sim \mathcal{U}([N])} \left[ - log \left( \frac{\textmyfont{exp}(z_i \cdot \overline{z}_{i0})}{\sum_{j \in [l]} \textmyfont{exp}(z_i \cdot \overline{z}_{ij})} \right) \right]
\end{align}
Here, it is the same $i$ that was being sampled in Equation \ref{eq:UserAttributionContrastiveLoss}.
Finally we take a convex combination of both the losses to get:
\begin{align}
    \label{eq:TotalUALoss}
    \mathcal{L}_{\text{RobustUA}} = \lambda\mathcal{L}_{\text{UA}} + (1 - \lambda)\mathcal{L}_{\text{aux}}
\end{align}
where, $0 < \lambda < 1$ is a hyper-parameter.
\paragraph{Note:} We do not have classes for a regression task. Therefore, we group the labels into $\mathcal{K}$ clusters by using K-means clustering where $\mathcal{K}$ is a hyper-parameter. Then we use the associated cluster labels as a proxy for the class labels. 

\subsection{\textit{Phase III}: Contextual Regularization (CR)}
\label{Subsection:Phase3}
Our primary assumption in this approach is that: A post is influenced by its context.
We demonstrate how to exploit the intuition ``hate begets hate'' \citep{10.1145/3415163} to our advantage.

\subsubsection{CR in classification}
\label{Subsubsection:ContextualClassification}
HateLM and UA, i.e., the methods described in section \ref{Subsection:Phase1} and \ref{Subsection:Phase2} (with the exception of Robust UA training) were self-supervised; hence, no information about the annotations available in the training set was used. After getting the pre-trained model in Phase II, we fine-tune using the annotations available for the dataset. We consider a pair of text sequence and its corresponding label $\langle{}\mathcal{S},\text{y}\rangle$ sampled from the training set uniformly at random, where $\text{y} \in [\mathcal{K}]$ denotes the true class label of the sampled sequence $\mathcal{S}$. Usually, we would get the vector embedding $z$ by passing $\mathcal{S}$ through the encoder. We would then have used the classifier in $\mathcal{M}$ to get the vector $v$ of $\mathcal{K}$ dimensions as follows:
\begin{align}
    \label{eq:ClassifierFunction}
    v = \mathcal{F}_\mathcal{C}(z; \theta_\mathcal{C})
\end{align}
where $\theta_\mathcal{C}$ parameterizes the classifier in $\mathcal{M}$.
According to the model, the probability of the sample belonging to $j^\text{th}$ class among $\mathcal{K}$ classes would be as follows:
\begin{align}
    \label{eq:ClassAssignmentProbabilityFromOutputs}
    \Prob[class(\mathcal{S})=j] = \frac{\textmyfont{exp}(v_j)}{\sum_{k\in[\mathcal{K}]}\textmyfont{exp}(v_k)}
\end{align}
and the most likely class assignment would be
\begin{align}
    \label{eq:MostLikelyClass}
    \hat{\text{y}} = \argmax_{j \in [\mathcal{K}]} \Prob[class(\mathcal{S})=j]
\end{align}
The cross-entropy loss would be calculated as:
\begin{align}
    \label{eq:ClassifierLoss}
    L_{\text{CE}} = \E_{i \sim \mathcal{U}([N])} \left[ - log \Prob[class(\mathcal{S})= \text{y}] \right]
\end{align}
We propose an additional method for regularization of the model using the contextual information during fine-tuning. For the given $\mathcal{S}$ we sample at most $n_a$ posts from the same thread (all comments/replies concerning the parent post) where the comment is posted (post context) and at most $n_b$ more posts from the timeline of the user who posted $\mathcal{S}$ (user context). Both post and user context is sampled without replacement.
So, we generate a context set $C$ of $\mathcal{S}$ for the $i^{\text{th}}$ text sequence sampled from the dataset:
\begin{align}
    \label{eq:ContextPostSet}
    C_i = \{ c_{i1}, ..., c_{in_a}, c_{i(n_a+1)}, ..., c_{i(n_a+n_b)} \}
\end{align}
We then generate the vector $v$ using $\mathcal{S}$ as mentioned in Equation \ref{eq:ClassifierFunction}. Next we generate $V_i$ from $C_i$:
\begin{align}
    \label{eq:ContextOutputSet}
    V_i = \{ v_{i1}, ..., v_{im} | \forall_{j \in [m]}v_{ij} = \mathcal{M}(c_{ij})\}
\end{align}
where $m=n_a+n_b$ and, $\mathcal{M}(.) \equiv \mathcal{F}_{\theta_\mathcal{C}}(\mathcal{F}_{\theta_\mathcal{E}}(.))$.
Our contextual loss from $V_i$ is as follows:
\begin{align}
    \label{eq:ContextOuterLoss}
    \mathcal{L}_{CCE,i} = \frac{-1}{m} \sum_{j \in [m]} log \left(
    \frac{\textmyfont{exp}(v_{ijt})}{\sum_{k \in [\mathcal{K}]} \textmyfont{exp}(v_{ijk})}
    \right)
\end{align}
where $t$ is the true class of the original labelled post. Therefore the auxiliary contextual cross-entropy loss is calculated as:
\begin{align}
    \label{eq:ContextAuxLoss}
    \mathcal{L}_{\text{CCE}} = \E_{i \sim \mathcal{U}([N])} \left[ \mathcal{L}_{CCE,i} \right]
\end{align}
Our final contextual regularization loss function is a linear combination of the losses mentioned in Equation \ref{eq:ClassifierLoss} and \ref{eq:ContextAuxLoss} as described below:
\begin{align}
    \label{eq:ContextualLoss}
    \mathcal{L}_{\text{Contextual}}^{\text{classification}} = \lambda\mathcal{L}_{\text{CE}} + (1-\lambda)\mathcal{L}_{\text{CCE}}
\end{align}
Back-propagating this loss we update the parameters $\theta_\mathcal{C}$ and $\theta_\mathcal{E}$, corresponding to the classifier and the encoder in $\mathcal{M}$ respectively, as one might have done normally.

\begin{table*}[ht]
    \centering
    \begin{tabular}{|l|c|c|c|c|}
        \hline
        \textbf{Dataset} & \textbf{Annotated} & \textbf{Total users} & \textbf{Labels} & \textbf{Max context} \\
        & \textbf{samples} &  &  & \textbf{sampled (per user)}\\
        \hline
        HateXplain & 10,000 & 3,300 & Hate speech, Offensive, Normal & 100 \\
        \hline
        LTI-GAB & 30,500 & 4,900 & Toxic and Non-Toxic & 50\\
        \hline
        Ruddit & 6,000 & 4,200 & Regression (-1 to +1) & 20 \\
        \hline
    \end{tabular}
    \caption{Dataset statistics for the datasets we used for performing the experiments along with the additional unsupervised data we collected. All datasets are in the English language.}
    \label{tab:dataset_stats}
\end{table*}

\vspace{-0.2cm}
\subsubsection{CR in regression}
\label{Subsubsection:ContextualRegression}
\vspace{-0.2cm}
The regression task is almost similar to classification as mentioned in section \ref{Subsubsection:ContextualClassification} with the exception that the $\text{y} \in \R$ in the tuple $\langle{}\mathcal{S},\text{y}\rangle$ sampled from the training set, and the output of the model $\mathcal{M}$ being a real value $r$ rather than a vector $v$ of $\mathcal{K}$ dimensions.
Since we use a regressor, Equation \ref{eq:ClassifierFunction} changes to
\begin{align}
    \label{eq:RegressorFunction}
    r = \mathcal{F}_\mathcal{R}(z; \theta_\mathcal{R})
\end{align}
where $\theta_\mathcal{R}$ parameterizes the regressor in $\mathcal{M}$.

The cross-entropy loss in Equation \ref{eq:ClassifierLoss} becomes squared loss as follows:
\begin{align}
    \label{eq:RegressorLoss}
    L_{\text{MSE}} = \E_{i \sim \mathcal{U}([N])} \left[ (\text{y}_i - r_i)^2 \right]
\end{align}
where $\text{y}_i$ is the label associated with the $i\text{th}$ example sampled from the training dataset, $r_i$ is the predicted value by model $\mathcal{M}$.

The techniques for selection of the context discussed in section \ref{Subsubsection:ContextualClassification} remain unaltered since our post and user context is unsupervised. However, the set of vectors $V_i$ becomes a set of real values $R_i$ generated as follows:
\begin{align}
    \label{eq:ContextOutputSet2}
    R_i = \{ r_{i1}, ..., r_{im} | \forall_{j \in [m]}r_{ij} = \mathcal{M}(c_{ij})\}
\end{align}
And our auxiliary contextual mean squared loss is calculated as the follows:
\begin{align}
    \label{eq:ContextRegressionLoss}
    \mathcal{L}_{CMSE} = \E_{i \sim \mathcal{U}([N])} \left[\frac{-1}{m} \sum_{j \in [m]} (\text{y}_i - r_{ij})^2 \right]
\end{align}
Our final contextual regularization loss function is thus a linear combination of the losses mentioned in Equation \ref{eq:RegressorLoss} and \ref{eq:ContextRegressionLoss} as described below:
\begin{align}
    \label{eq:ContextualRegressionLoss}
    \mathcal{L}_{\text{Contextual}}^{\text{regression}} = \lambda\mathcal{L}_{\text{MSE}} + (1-\lambda)\mathcal{L}_{\text{CMSE}}
\end{align}


\vspace{-0.12cm}

\begin{table*}[t!]
    \centering
    \small
    \begin{tabular}{l|l|cc|cc|cc}
        \toprule
        \multirow{2}{*}{\textbf{Phase}} & \multirow{2}{*}{\textbf{Model}} & \multicolumn{2}{c|}{\textbf{HX-GAB}\footnotemark[5] (3 class)} & \multicolumn{2}{c|}{\textbf{LTI-GAB} (2 class)} & \multicolumn{2}{c}{\textbf{Ruddit} (regression)} \\ \cline{3-8}
        &  & Acc & Macro-F1 & Acc & F1 (toxic) & MSE & MAE \\
        \midrule
        \midrule
        Existing approaches & TF-IDF & 0.6337 & 0.5604 & 0.8993 & 0.8824 & 0.1128 & 0.2651 \\
        as baselines & BERT & 0.6763 & 0.6376 & 0.9141 & 0.9019 & 0.1041 & \textbf{0.2521} \\ 
        & HateXplain$^\dagger$\footnotemark[5] & \textbf{0.6905} & \textbf{0.6511} & \textbf{0.9177} & \textbf{0.9062} & \textbf{0.1036} & \textbf{0.2520} \\
        \midrule
        Continual pre-train- & HateBERT & 0.6843 & 0.6458 & 0.9166 & 0.9040 & 0.1029 & 0.2516 \\ 
        ing (CP phase) & HateLM & \textbf{0.6882} & \textbf{0.6493} & \textbf{0.9174} & \textbf{0.9058} & \textbf{0.1018} & \textbf{0.2474} \\
        \midrule
        User Anchored self- & BERT + UA & 0.6950 & 0.6632 & 0.9192 & 0.9089 & 0.0994 & 0.2367 \\ 
        supervision (UA phase) & HateXplain + UA & 0.7058 & 0.6680 & 0.9211 & 0.9105 & 0.0989 & 0.2329 \\
        & HateLM + UA & \textbf{0.7087} & \textbf{0.6711} & \textbf{0.9215} & \textbf{0.9117} & \textbf{0.0958} & \textbf{0.2229} \\
        \midrule
        Contextual Regu- & BERT + CR & 0.6819 & 0.6452 & 0.9176 & 0.9051 & 0.1019 & 0.2438 \\ 
        larization (CR phase) & HateXplain + CR & \textbf{0.6935} & 0.6555 & \textbf{0.9199} & 0.9088 & 0.1011 & 0.2410 \\ 
        & HateLM + CR & 0.6924 & \textbf{0.6558} & \textbf{0.9200} & \textbf{0.9096} & \textbf{0.0995} & \textbf{0.2342} \\
        \midrule
        UA and CR phase & BERT + UA + CR & 0.7017 & 0.6673 & 0.9211 & 0.9104 & 0.0968 & 0.2314 \\ 
        together & HateXplain + UA + CR & 0.7099 & 0.6702 & 0.9236 & 0.9122 & 0.0955 & 0.2291 \\ 
        & CRUSH$^\divideontimes$ & \textbf{\textit{0.7133}} & \textbf{\textit{0.6749}} & \textbf{\textit{0.9234}} & \textbf{\textit{0.9149}} & \textbf{\textit{0.0921}} & \textbf{\textit{0.2188}} \\
        \bottomrule
    \end{tabular}
    \caption{
        \textbf{Experimental outcomes of our approaches.} Our model \textbf{CRUSH} $\equiv$ HateLM + UA + CR. For reporting results (except CR and UA+CR phases) the encoder models were attached with a classifier (refer section \ref{Subsection:Implementation}) and fine-tuned (using the loss mentioned in Equation \ref{eq:ClassifierLoss}). \textit{Across the columns}: we show the results with three datasets. In HateXplain (HX)'s and Learning to Intervene (LTI)'s results, higher metrics are better. In Ruddit results, lower metrics are better. \textit{Across the rows}: we show the various models grouped by the \textit{phase} of training along with corresponding baselines. In each group, we have indicated the best-performing models' results corresponding to the evaluation metric in bold. The overall best performing models' results have additionally been italicized. \textbf{The models denoted by $\dagger$ (the best competing baseline) and $\divideontimes$ (CRUSH) are significantly different (M-W U test with $p<0.05$) across all datasets.}}
    \label{table:MainResults}
\end{table*}

\section{Experiments}
\label{Section:Experiments}
\vspace{-0.12cm}
We experiment with the downstream task of hate speech detection. We establish the effectiveness of our proposed approaches by demonstrating that they are:
\textbf{(i)} mutually independent of each other,
\textbf{(ii)} independent of the base language model used, and
\textbf{(iii)} applicable across various social network datasets. We validate these claims by comparing our approaches with the following baselines:\\
\textbf{(i)} To establish mutual independence among the training phases, we train a base language model (BERT from \citet{devlin-etal-2019-bert}) with each of the training phases separately. An improvement over the performance of the base model in each phase would indicate the advantage of our approach over na\"ive training. \\
\textbf{(ii)} To establish independence from the base language model, we train multiple baseline language models during each of the above-mentioned separate training phases (HateBERT during Phase I; BERT and HateXplain during the remaining phases). Consistent performance improvement irrespective of the language model would again indicate the advantage of our approach.\\
\textbf{(iii)} We use three datasets from two different social networks and two types of tasks (classification and regression). This establishes the capability of our methods to solve a range of heterogeneous tasks across contrasting datasets.

We next describe the datasets and the implementation details of our approaches.
\subsection{Datasets}
\label{Subsection:Datasets}
\noindent\textbf{Social networks}: In particular we experiment with two popular social networks (a) \url{gab.ai} (GAB), and (b) \url{reddit.com} (Reddit).
Our choice of the dataset was guided by the availability of the context graph and additional data collection time. \citet{JasonPushshift} had scraped GAB and made the network freely available for academic use. On the other hand, Reddit has an API\footnote[4]{\hyperlink{https://www.reddit.com/dev/api/}{https://www.reddit.com/dev/api/}} available to get the public domain data. Therefore, these websites were favorable for our experiments. Since Reddit involved additional data collection (a time consuming process), we chose a popular dataset that contains less than 10,000 datapoints. 

\vspace{0.1cm}
\noindent\textbf{Annotated hate speech data}:
We use the following english hate speech datasets for our experiments (See Table \ref{tab:dataset_stats} for more information on dataset statistics) --
(i) HateXplain-GAB dataset~\cite{Mathew_Saha_Yimam_Biemann_Goyal_Mukherjee_2021} (contains data from GAB), (ii) LTI-GAB dataset~\cite{qian-etal-2019-benchmark} (contains data from GAB) and, (iii) Ruddit~\cite{hada-etal-2021-ruddit} (contains data from Reddit).

\begin{figure}[ht]
    \centering
    \includegraphics[width=0.9\linewidth]{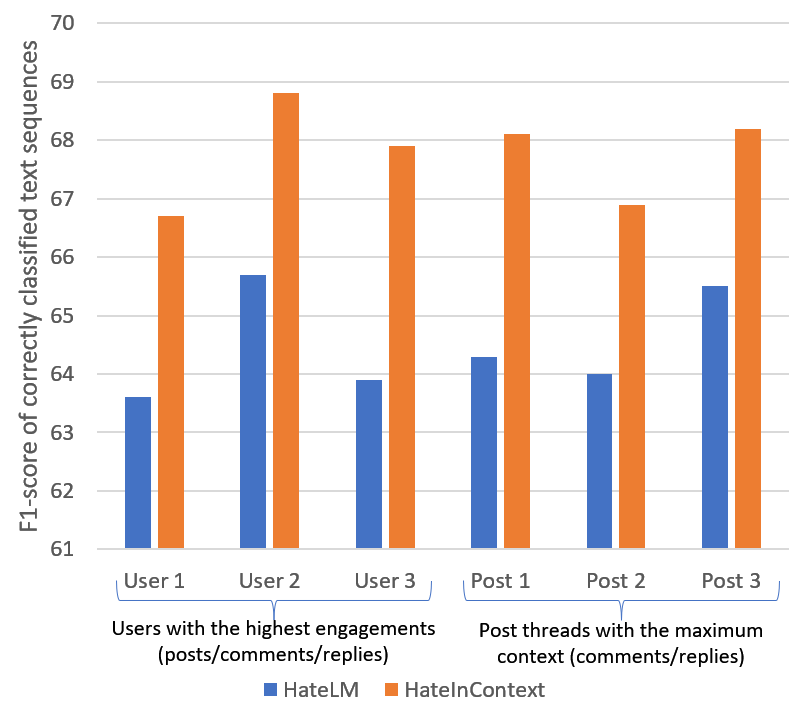}
    \caption{[Best viewed in color] F1-scores of correctly classified posts grouped by three user contexts and three post contexts. The improvement along the user contexts and post contexts demonstrate the validity of the UA and CR phases of CRUSH training respectively.}
    \label{fig:model_component_ablation}
\end{figure}
We use \textit{only} the GAB subset of the annotated data from both the datasets (i) and (ii), because the social network context graph for GAB is publicly available. The GAB subset of dataset (i) has around $10$K annotated data samples, which are already divided into $80$-$10$-$10$ train-val-test split. Dataset (i) has three class annotations (hate speech, offensive, and normal).
Dataset (ii) contains intervention sentences along with which sentences to intervene and a binary label - hate and non-hate.
We use a $90$-$10$ train-test split with the random seed $2021$, and among the training set we use $10\%$ randomly sampled data for validation.
Dataset (iii) was collected from Reddit and is labeled for regression task. The dataset contains ratings corresponding to the measure of hate within a sentence ranging between $-1$ and $1$, with higher numbers corresponding to a higher extent of hate. There are about $6$K examples in this dataset. We use a train test split similar to that we used in the dataset (ii).

Our pre-processing procedure for all the textual data, both labeled and unlabelled is exactly the same as that of HateXplain. \cite{Mathew_Saha_Yimam_Biemann_Goyal_Mukherjee_2021}.

\footnotetext[5]{HateXplain refers to both a dataset and the corresponding model provided in the paper. However, since we only use the GAB subset of the HateXplain dataset, we require to fine-tune the HateXplain model on this subset, for it to be comparable to other models, and therefore it is considered a baseline.}

\vspace{0.1cm}
\noindent\textbf{Unlabelled data for self-supervision}:
We get the threads corresponding to the annotated Reddit datapoints using the Reddit API and use that as our unlabelled corpora for self-supervision in case of the regression task.
For GAB, we use the whole network datadump available on Pushshift \cite{JasonPushshift} as mentioned in Section \ref{Subsection:Phase2}.


\begin{figure*}[!t]
	\centering
	\begin{subfigure}[t]{0.32\textwidth}
		\centering
		\includegraphics[width=\linewidth]{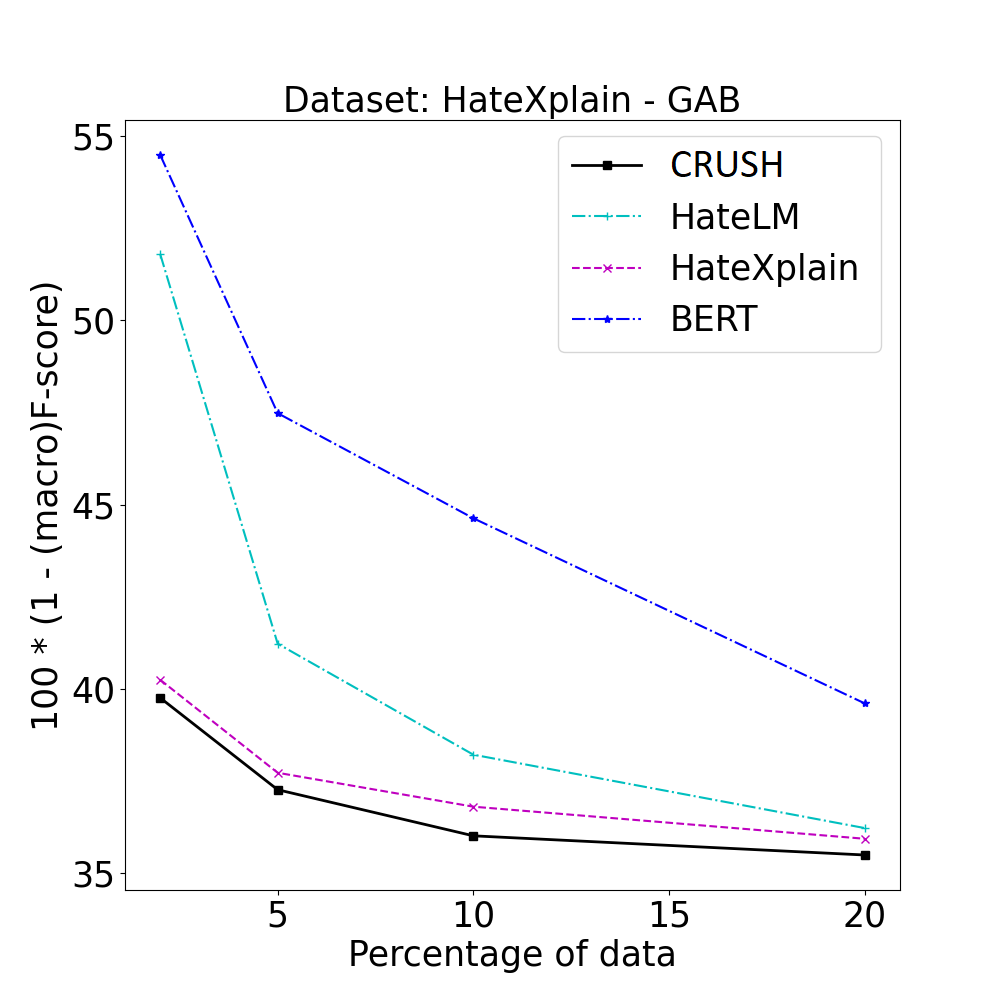}
	\end{subfigure}
	\begin{subfigure}[t]{0.32\textwidth}
		\centering
		\includegraphics[width=\linewidth]{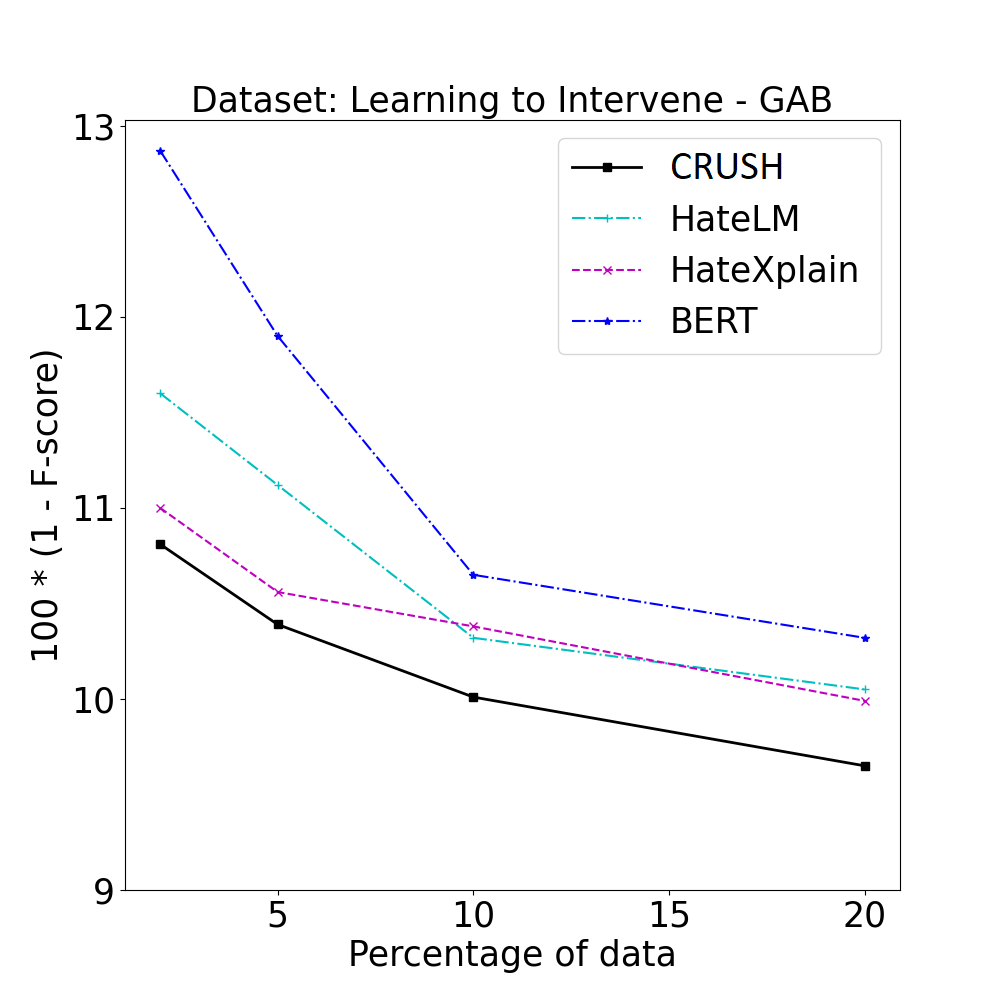}
	\end{subfigure}
	\begin{subfigure}[t]{0.32\textwidth}
		\centering
		\includegraphics[width=\linewidth]{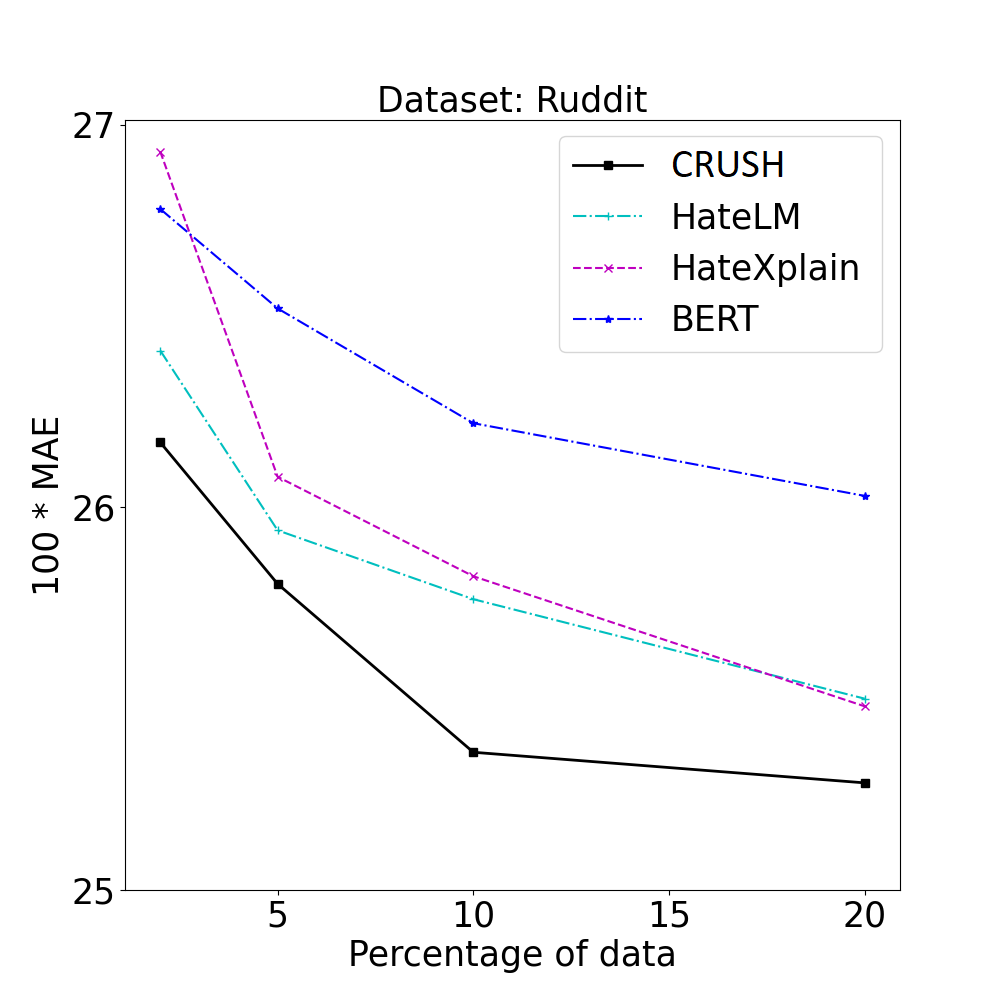}
	\end{subfigure}
	\caption{[Best viewed in color] Comparison of model error in low data regime across the three datasets. Our model (black solid line) consistently outperforms existing approaches that do not use the social network context.  
	For plots showing comparisons with non-contextualized LMs (e.g., TF-IDF) as well, refer to Figure \ref{full_graphs_ablation} in Appendix.
	}
	\label{magnified_graphs_ablation}
\end{figure*}
\subsection{Implementation details}
\label{Subsection:Implementation}

For our encoder, we use the pre-trained \textmyfont{bert-base-uncased} that is available on huggingface \footnote[6]{\hyperlink{https://huggingface.co/bert-base-uncased}{https://huggingface.co/bert-base-uncased}}. 
\begin{table}[ht]
    \centering
    \begin{tabular}{|l|l|l|}
        \hline
        \textbf{Word} & \textbf{BERT} & \textbf{HateLM}  \\
        \hline
        black & everywhere, & racist, evil,  \\
        \hline
        & free, not & stupid \\
        muslim & accepted, opti- & terrorist, \\
        & onal, allowed & evil, shit \\
        \hline
        jews & persecuted, ex- & idiots, bad, \\
        \hline
        & cluded, present & stupid \\
        men & equal, free, & evil, people, \\ 
        & citizens & bad \\
        \hline
        women & excluded,  & dumb, evil, \\
        \hline
        & only, bias & stupid \\
        \hline
    \end{tabular}
    \caption{Top-3 next word predictions by each LM when the corresponding word is given as prompt. The hateful words as the output of the next word prediction demonstrate the effectiveness of the hate-infusion into LM.}
    \label{table:HateInfusionAblation}
\end{table}
By continual pre-training on this model, we obtain our HateLM using the text sequences from GAB and RAL-E as described in section \ref{Subsection:Phase1}.
Afterward, we can obtain a $768$ dimensional pooled output from our encoder which can then be passed onto the classifier/regressor.
For both our classifier and regressor model, we select a neural network consisting of a $2$ Fully-connected layers. The hidden layer for the same is chosen to have $128$ dimensions along with a ReLU activation. The input dimension is dependent on the output of the encoder, and in our case is set to $768$.
The output dimension is $1$, $2$ or $3$ corresponding to the regressor, binary and ternary classifier. We used \textit{Adam} optimizer with batch size $48$, max seq length $128$ and learning rates $3e$-$6$ and $2e$-$5$ for our encoder and classifier/regressor respectively.
Our approaches do not require additional context during inference (see Figure \ref{img:ModelDiagrams}(d)).

\vspace{-0.12cm}
\subsection{Ablation studies}
\vspace{-0.12cm}
We frame the ablation studies to answer a series of interesting questions. These are:\\
\textbf{Q1}: How much does self-supervised learning objective during pre-training help in the downstream task (if at all)? \\
\textbf{Q2}: Similarly, how much does contextual regularization during fine-tuning help? \\
\textbf{Q3}: Do these two approaches work only in conjunction with each other?

In addition to the above, we perform the following ablation experiments to validate the usefulness of various components of our model:\\
\textbf{Q4}: How much of the performance can be attributed to the detection of isolated hateful words without any contextual information in place? To answer this question, we use the TF-IDF \cite{tfidf} vector embedding of each text sequence. since TF-IDF computes representations solely based on the frequency of words without any contextual information. \\
\textbf{Q5}: Have the hateful words been contextually incorporated into the hate-infused language model? Here, we prompt our hate-infused language model and study the next word it predicts. \\
\textbf{Q6}: How do these approaches perform in a few-shot setting? 
\vspace{-0.12cm}



\section{Results}
\label{Section:Results}

The Table \ref{table:MainResults} compares the performance of our methods with related models and baselines. We report the performance metrics according to the test sets of the corresponding datasets.
\vspace{-0.2cm}
\paragraph{[1]}
It is evident from Table \ref{table:MainResults} that both CP and UA phases (self-supervised phases) lead to improvements over the baselines. This answers \textbf{Q1}. To answer how useful regularization was we can look at the CR phase results which are again better than the baselines thus answering \textbf{Q2}. 
\begin{table*}[ht]
    \centering
    \begin{tabular}{|l|c|c|c|}
        \hline
        \textbf{Post (along with ground-truth label)} & \textbf{BERT} & \textbf{HateLM} & \textbf{CRUSH} \\
        \hline
        if your humor is based on racism homophobia sexism and rape & \xmark & \cmark & \cmark \\
        you're not f**king funny go home (\textbf{non-hate}) & & & \\
        \hline
        <user> ah man i f**king hate you so much (\textbf{hateful}) & \xmark & \cmark & \cmark \\
        \hline
        the french government position is that france is made stronger by & \xmark & \xmark & \cmark \\
        the immigrants it is bribing to leave porter (\textbf{non-hate}) &  &  &  \\
        \hline
        i am presuming he means the standard left wing idiots lots of & \xmark & \xmark & \cmark \\
        hypocritical women amongst them think maria ladenburger obvi- &  &  & \\
        ously you are against these muhammedans (\textbf{hateful}) &  &  &  \\
        \hline
        a covington catholic a native american indian and a black hebrew & \xmark & \xmark & \xmark \\
        israelite walk into a bar (\textbf{non-hate}) & & & \\
        \hline
        if money was grown on trees women would be dating monkeys & \xmark & \xmark & \xmark \\
        oh wait never mind (\textbf{hateful}) &  &  & \\
        \hline
    \end{tabular}
    \caption{Qualitative results using text sequences from HateXplain dataset. \cmark \ \  indicates sentences were classified properly by the corresponding models while \xmark \ \  indicates incorrect classification. MLM based training of BERT and HateLM does not seem to capture the complexities of hate speech properly, thus CRUSH outperforms them.} 
    \label{tab:quali_results}
\end{table*}

Further, we can also see that UA and CR phase results individually beat the baseline. Furthermore, combined UA+CR phase outperforms all models. Hence, the conjunction hypothesis is valid thus answering \textbf{Q3}.
\vspace{-0.125cm}
\paragraph{[2]}\textbf{Q4} is answered by the considerable improvement over the TF-IDF baseline as can be seen in Table \ref{table:MainResults}. In addition, Figure \ref{fig:model_component_ablation} shows the advantages gained by using CRUSH over HateLM. It demonstrates both the user-level and the post-level contextual information from the network incorporated into our CRUSH model are individually helpful over models that do not have social network context information.
\vspace{-0.125cm}
\paragraph{[3]}Table \ref{table:HateInfusionAblation} answers \textbf{Q5} by showing some of the prompt completion results for BERT vs our Hate-infused LM. It can be noticed that although HateLM definitely has been incorporated with racial and ethnic hate, it surprisingly does not discriminate between genders.
\vspace{-0.125cm}
\paragraph{[4]}Figure \ref{magnified_graphs_ablation} shows the few-shot training results with small percentages of data sampled from our training datasets (2\% to 20\% data points) uniformly at random. Intuitively our model is already able to outperform all other baselines consistently even without context because of the self-supervised training procedures we propose during the CP and UA phases which captures the hate \& users' style+bias. This is evident from the results of CP \& UA phases in Table \ref{table:HateInfusionAblation}. So later adding context in few-shot setting simply further enhances our model thus answering \textbf{Q6} (see Appendix for full results).

\vspace{0.125cm}
\noindent\textit{Discussion of some qualitative examples}:
Table \ref{tab:quali_results} presents a few text sequences from the HateXplain dataset and their corresponding results (classification/missclassification) for the models -- BERT, HateLM, and CRUSH.
It can be noticed that the first two sentences are properly classified by HateLM and CRUSH but not by vanilla BERT. That leads us to believe that BERT predictions are heavily dependent on the token occurrences rather than the context in the sentence where the tokens have occurred.
As for the next two sentences, it is evident that the sentence representations learnt by CRUSH (due to UA + CR phases of training, see Section \ref{Subsection:Phase2}, \ref{Subsection:Phase3}) are superior to those learnt by the HateLM, hence it classifies these text sequences better than simple MLM based training.
Finally, the last two sentences indicate that CRUSH (along with the other approaches) still lacks the ability to identify humor, sarcasm, and implicit hate speech, which are known to be difficult problems.



\vspace{-0.2cm}
\section{Conclusion}
\label{Section:Conclusion}
In this paper, we provide approaches to infuse social network context using the self-supervised user-attribution pre-training objective combined with the contextual regularization objective on top of traditional MLM for hate speech tasks. We empirically demonstrate the advantage of our methods by improving over existing competitive baselines in hate speech detection and scoring tasks across three different datasets. We also show that our method performs superior in the low data regime as well when compared to existing approaches. We also do ablations to understand the benefits of each objective separately.
Future work include exploiting the relations among users, using different base models capable of incorporating longer contexts, and trying to address hard problems like sarcasm and implicit hate speech detection in social networks.


\section{Ethical considerations}

All the datasets that we use are publicly available. We report only aggregated results in the paper. For context mining, we have used data either available in the public domain or available through official APIs of the corresponding social media. Neither have we, nor do we intend to share any personally identifiable information with this paper. We also make our codebase publicly available here - \url{https://github.com/parag1604/CRUSH}.

Our model CRUSH helps advance the state-of-the-art in hate speech detection by incorporating continual pre-training, capturing user writing biases and leveraging both user \& post context (which is publicly available as well). This in turn should be able limit the amount of hateful threads in social networks/media by better detection of hate speech, thus promoting a more friendly and welcoming environment for people from all race, religion, ethnicity, gender etc.

Unfortunately, these models are not completely free from all potential negative impact. One such example being that our models have hate knowledge infused within them during the CP phase (refer Section~\ref{Subsection:Phase1}) of our training pipeline. As shown in Table~\ref{table:HateInfusionAblation}, these language models could be used potentially used for generating hateful words/sentences given an initial prompt.

However, the hateful words generated by the HateLM can be also identified by the platform if the platform uses a hate speech detection algorithm like ours or uses a set of hate lexicons to directly filter out such keywords generated \cite{Gitari2015ALA}. Moreover, our final model - CRUSH - is not exactly suitable for toxic language generation as it is further pre-trained on the user style discrimination task and fine-tuned on the hate speech classification task making its classification capabilities (potential use by the social media platforms to detect hate speech) stronger than the hate speech generation capabilities (the potential of the model being abused with malicious intent). So, the platforms using our better-informed model will easily detect any hate speech generated by the model.

Moreover, large-scale technologies for better generation of hate speech using GPT-2 and other generative models \cite{wullach2021fight,hartvigsen2022toxigen} (with the intention to train hate speech classifiers better) already exist and as a model trained only for classification, our model is likely to be far weaker than these models for generation of harmful content. While these language models can be used both to benefit or disrupt our lifestyle just like any other technology \cite{dual}, we urge the researchers to exercise ultimate caution while using them, including ours. For the same reason, we do not make the trained model parameters publicly available (except for the code to promote reproducibility).  

Also, one of the factors increasing bias in the classification model is the data it is trained upon. Hence, any potential bias in the datasets that are manually annotated by human beings (who are not free from bias) can result the model being biased for/against some specific target groups.

To incorporate better inductive bias into the model, we have trained the model to initially map text sequences posted by the same user to vectors that are close in the embedding space. This helps the model to identify the dialects of the various users and help the model perform better. However, this inductive bias may cluster the linguistic characteristics of some groups which in turn might potentially increase the chances of that particular language style being classified as hate speech if the annotated data contains only hateful instances from that particular dialect. This may make the model biased if those particular dialects are used predominantly by some protected category (Blacks, Jews, Women, Mexican etc.).

The most simple and effective solution here would be to annotate data for each dialect/vernacular group, potentially stratifying the dataset into clusters using the pre-trained language model. If there are balanced examples from each of these dialect/vernacular or at least each protected category (Blacks, Jews, Women, Mexican etc.) for both the hate and non-hate categories, such biases can be well mitigated.

Further, our model like any other hate speech model is not suitable for in-the-wild deployment without explicit human scrutiny. Language use is often very specific to each platform. So, the distribution of words and data-points may not match the training data distribution of the model. Thus, it is extremely important to first test the model on the platform, check for potential biases against the protected categories and individual linguistic groups/dialects and deploy it as a preliminary filter assisting the human experts detecting hate speech.

\section{Acknowledgements}
We thank Facebook (Meta Platforms, Inc.) for sponsoring the stay of PD at IIT Kharagpur through the Ethics in AI Research grant and Tata Consultancy Services (TCS) for funding SC with the TCS PhD fellowship.


\bibliography{custom}
\bibliographystyle{acl_natbib}

\newpage

\onecolumn
\appendix
\label{sec:appendix}
    \section{\Large Appendix}



\begin{figure*}[ht]
	\centering
	\begin{subfigure}[t]{0.49\textwidth}
		\centering
		\includegraphics[width=\linewidth]{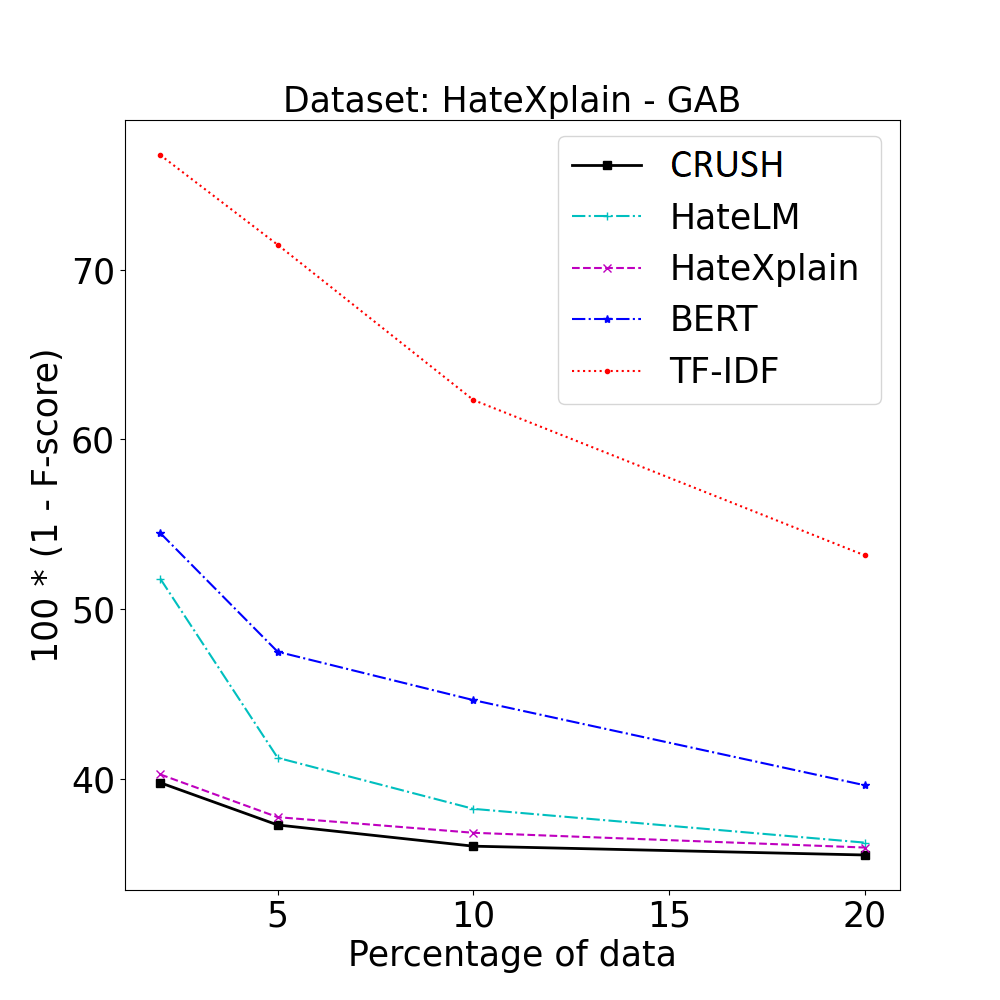}
	\end{subfigure}
	~
	\begin{subfigure}[t]{0.49\textwidth}
		\centering
		\includegraphics[width=\linewidth]{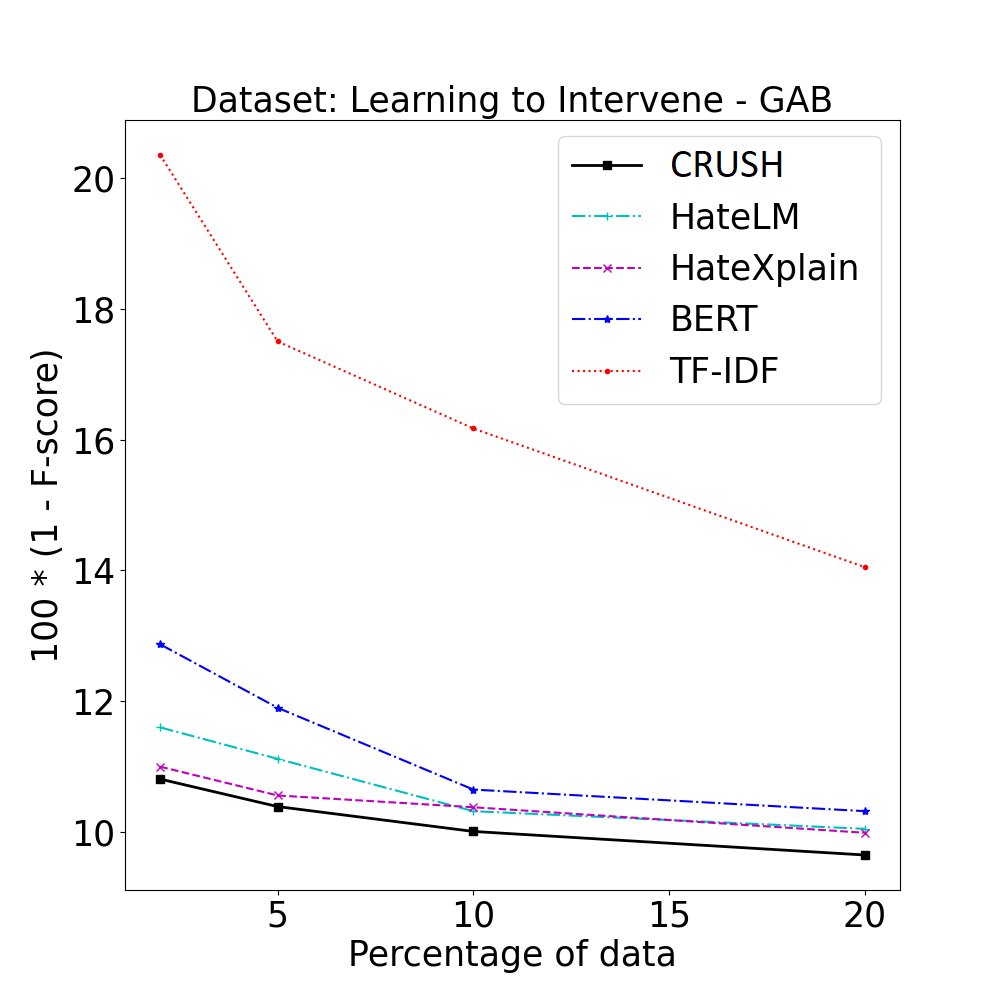}
	\end{subfigure}
	\\
	\begin{subfigure}[t]{0.49\textwidth}
		\centering
		\includegraphics[width=\linewidth]{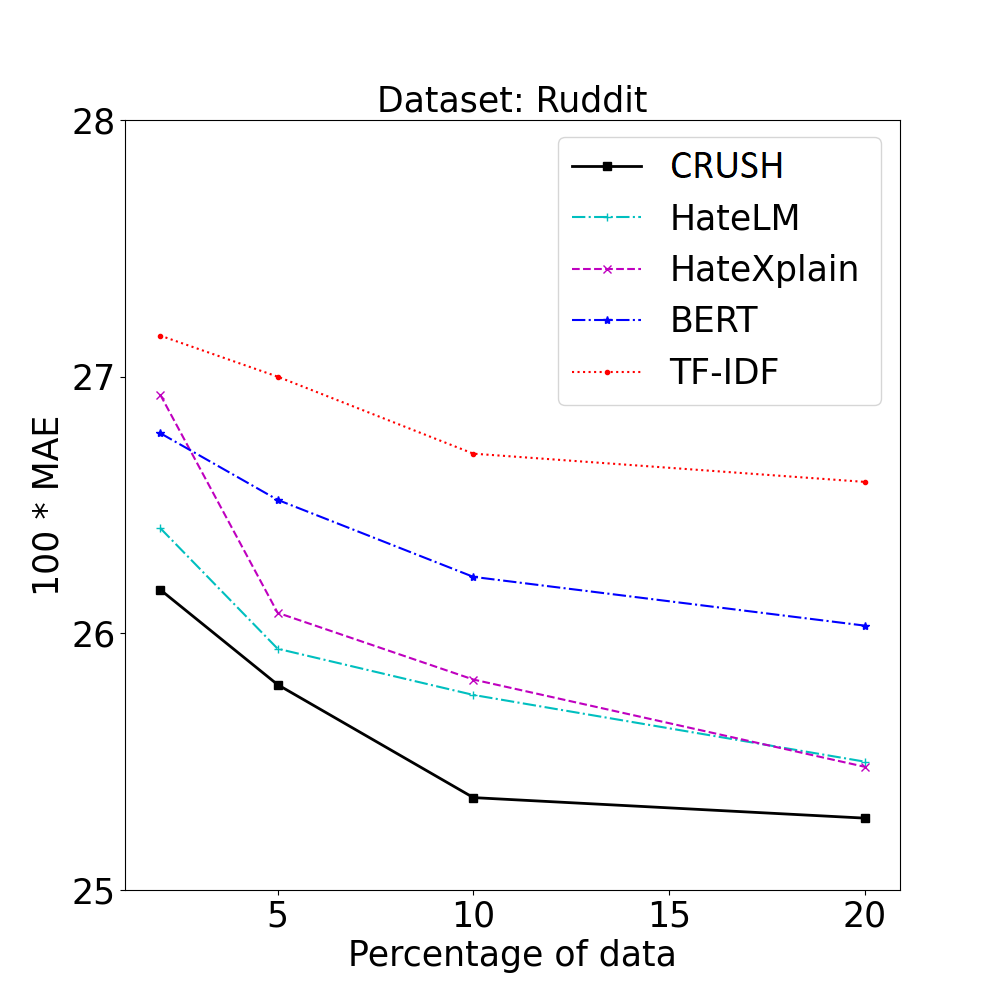}
	\end{subfigure}
	\caption{Full comparison of model errors in low data regime. This includes a comparison of our CRUSH model with baseline models built on non-contextualized and contextualized embeddings.}
	\label{full_graphs_ablation}
\end{figure*}


\end{document}